# Model-based active learning to detect an isometric deformable object in the wild with a deep architecture

**Shrinivasan Sankar**     **Adrien Bartoli**

Université Clermont Auvergne, France

Corresponding author: Shrinivasan Sankar

`vasan.shrini@gmail.com`

June 7, 2018

**Abstract**

In the recent past, algorithms based on Convolutional Neural Networks (CNNs) have achieved significant milestones in object recognition. With large examples of each object class, standard datasets train well for inter-class variability. However, gathering sufficient data to train for a particular instance of an object within a class is impractical. Furthermore, quantitatively assessing the imaging conditions for each image in a given dataset is not feasible. By generating sufficient images with known imaging conditions, we study to what extent CNNs can cope with hard imaging conditions for instance-level recognition in an active learning regime.

Leveraging powerful rendering techniques to achieve instance-level detection, we present results of training three state-of-the-art object detection algorithms namely, Fast R-CNN, Faster R-CNN and YOLO9000, for hard imaging conditions imposed into the scene by rendering. Our extensive experiments produce a mean Average Precision score of 0.92 on synthetic images and 0.83 on real images using the best performing Faster R-CNN. We show for the first time how well detection algorithms based on deep architectures fare for each hard imaging condition studied.

# Contents

# 1 Introduction

Object detection comes seamlessly for humans but is a hard and challenging problem for computers. Over the last three years, aided by the availability of powerful computing machinery, object detection has seen immense progress both in terms of accuracy and speed. By and large, this can be attributed to the vast availability of data and a variation of Neural Networks for images, the CNNs. CNNs are specialized Neural Networks that place constraints on weights to learn invariances in high dimensional inputs such as images (LeCun et al., 1989, 1998). They have been extremely successful recently for a range of computer vision tasks such as image classification, object detection and semantic segmentation. Object detection is challenging mainly due to the possibility of variations in object appearance. Objects may look very different in different poses and illuminations; deformations also lead to large intra-class variations, thereby adding another level of complexity to detection. Yet, the influence of hard imaging conditions on the CNN based detection algorithms remains largely unexplored. Detection algorithms need to tune themselves to cope with both deformations and hard imaging conditions as and when they occur. A common approach to improve the performance of CNN based models is data augmentation (Krizhevsky et al., 2012). However, possibilities are limited in data augmentation (flipping, cropping, etc.). For instance, pose changes and deformations of an object are not augmentable. An alternative to sidestep this limitation could be to use powerful rendering techniques and generate synthetic data 'on demand' as and when failure cases occur at test time. Of course, bounding box annotations come for free with rendering. Most standard detection datasets like PASCAL VOC come with large examples of *category-level* annotations for rigid objects such as cars, bicycles and chairs. However, gathering sufficient data at the *instance-level* for supervised learning is much more challenging and impractical. Our proposed method generates sufficient data to feed the data hungry CNNs for instance-level recognition.

We set out to research the possibility of using synthetic data generated by rendering to train a CNN based object detection algorithms by an active learning approach to achieve instance-level detection. To this end, we propose to detect deformable objects occurring in natural scenes under 'in the wild' imaging conditions. More specifically, we focus on developable surfaces undergoing isometric deformations. Perriollat *et al* introduced a way to parameterize the deformations of a developable surface based on guiding rules and their bending angles (Perriollat and Bartoli, 2013). We use their paper model toolbox to generate random deformations of a developable surface and render them with backgrounds in order to generate our training and test datasets. In short, our contributions are as follows. ($i$) We show to what extent CNNs can learn and cope with different imaging conditions. We dub this the *learnability* of imaging conditions. ($ii$) With active learning for deep learning remaining a largely unexplored area, we propose an active learning algorithm that can generate images of failure conditions in order to actively adapt the trained model on-the-fly. (*iii*) We release an annotated object detection dataset with five deformable thin-shell objects.

We show extensive experimental results using three state-of-the-art object detection algorithms: Fast R-CNN (Girshick, 2015), Faster R-CNN (Ren et al., 2015) and YOLO9000 (Redmon and Farhadi, 2016) which are the baseline methods considered throughout our work. The active learning algorithm we propose achieves a mean Average Precision (mAP) score of 0.92 on synthetic images. By fine tuning the model trained on synthetic images with as little as 100 real images, we obtain a mAP score of 0.83 on the best performing Faster R-CNN algorithm.

## 2   Related Work

We review the available literature on object detection using local features, object detection using CNNs, the use of object proposals for object detection and active learning in deep learning.

### 2.1   Object Detection with Local Features

Object detection with local features works by extracting a set of keypoints from both the target object model and the test image. Local invariant descriptors computed in regions around the keypoints from both the target object and the test images are compared to establish matches and hence detect the presence of the target object. In this review, we restrict ourselves to deformable object detection. A real-time detection method using a fast wide-baseline matching algorithm was proposed in (Pilet et al., 2008). Pizarro *et al* showed that occluded regions in deformable objects can be reasoned out and hence detected (Pizarro and Bartoli, 2012). Based on (Pizarro and Bartoli, 2012), Alcantarilla *et al* detected deformable objects in a database by extracting and storing SURF feature descriptors for each object in the database and achieved deformable 3D reconstruction (Alcantarilla and Bartoli, 2012). The feature descriptors of a given test image were compared to those in the object database. The detection method returned a set of potential objects, each of which is then validated or eliminated by the 3D reconstruction step. (Shaji et al., 2010) proposed a method to simultaneously solve for correspondences and 3D shape. (Sánchez-Riera et al., 2010) used keypoints to estimate the camera pose in addition to estimating the 3D shape and correspondences. (Tran et al., 2012) showed that even simple schemes such as RANSAC can be effective tools for outlier rejection in deformable object registration. Though these methods are effective in some controlled imaging conditions, they rely heavily on keypoints which are not guaranteed to be found under hard imaging conditions such as blur.

## 2.2  Object Detection with CNNs

The idea of using CNNs for image classification is quite mature and dates back to the 1990s when the CNNs were first introduced (LeCun et al., 1998). However, using CNNs for object detection springs from the introduction of Region-based CNN (Girshick et al., 2016) alongside Overfeat (Sermanet et al., 2013) and (Erhan et al., 2014). The idea in (Girshick et al., 2016) is to first generate bottom up proposals using out of the box object proposal systems. The different regions generated are warped to a fixed size in order to be fed to a CNN classifier trained for image classification. Class specific Support Vector Machines (SVMs) at the output of the CNN serve to identify the proposal bounding boxes as positives. Bounding box regression is then used to further narrow down the obtained bounding boxes. The main drawback of R-CNNs is that it takes 10 to 45 seconds per image during test time. Fast R-CNN (Girshick, 2015), as the name suggests, addresses the speed problem of R-CNN and improves it to near real-time. Furthermore, it overcomes the multi-stage pipeline of R-CNN by squeezing bounding box regression and classification together. Fast R-CNN introduces two changes: 1) Region of Interest (RoI) pooling after the feature extracting layers (convolution and pooling). The result of RoI pooling is a vector of fixed size which is fed through the fully-connected layers. 2) Computing class probabilities and bounding box locations jointly. This consolidates the bounding box regression step of R-CNN into the network itself. Both Fast R-CNN and R-CNN have external proposal systems clearly limiting their speed to that of the proposal system chosen. Additionally, each proposal has to be validated by the network. Faster R-CNN proposes an elegant solution by embedding detection proposals into the network and calling it *Region Proposal Network* (Ren et al., 2015). The authors of (Ren et al., 2015) argue that proposals can be generated from feature maps (convolution layer outputs) rather than images. They fix 'anchor boxes' at reference points on feature maps and sample them at different scales and aspect ratios. The unified network (region proposal + detection) takes detection to 5fps even with very deep network architectures. You Only Look Once (YOLO) is the first end-to-end CNN detector proposed. It is based on dividing the input image into grids and letting the grid falling at the centre of an object responsibly estimate the bounding box location and confidence score for the object. If a grid does not contain an object, it should have a confidence score of zero. Thus it poses detection as a regression problem (Redmon et al., 2016). Though YOLO is one of the fastest object detector available, it compromises performance for speed. YOLO9000 is an improved version of YOLO that harnesses the vast availability of classification data to improve detections (Redmon and Farhadi, 2016). In our work, we adapt three of the above mentioned state-of-the-art detectors: Fast R-CNN, Faster R-CNN and YOLO9000 for the active learning and learnability algorithms we propose.

## 2.3  Object Proposals

Seminal works on proposals started from *Objectness* (Alexe et al., 2010, 2012). In Objectness, each candidate window receives a score based on three factors: well defined closed boundaries, how uniquely the object stands out in an image and distinction of the object from its surroundings. Selective search (Uijlings et al., 2013) works by bottom-up grouping of regions. For identifying initial regions, it uses graph-based segmentation. The over-segmented regions are then combined together based on the similarity of each region to its neighbors. The most similar regions are grouped and the process continues until the entire image is covered. *Edge Boxes* works by computing edge responses for each pixel using Structured Edge Detector (Dollar and Zitnick, 2013). Edge peaks are computed from edge responses leading to each pixel having a magnitude and orientation. Manen *et al* proposed the Prime object proposals based on the randomized Prime algorithm (Manen et al., 2013). It works by generating connectivity graphs between image superpixels and weighing them with the probabilities of superpixels belonging to the same object. Binarised Normed Gradients (BING) was one of the fastest proposal systems introduced that works based on the fact that the norm of gradient are strongly correlated for objects with tightly closed boundaries (Cheng et al., 2014). BING trades-off performance for speed. Edge Boxes on the other hand runs fast without compromising on performance (Hosang et al., 2014). Furthermore, Edge Boxes provides explicit parameters to tune the number of proposals (Zitnick and Dollár, 2014). For this reason, we chose Edge Boxes over others.

## 2.4  Active Learning for Deep Learning

Active Learning (AL) has been a long studied area of Machine Learning. The key idea being that a learning algorithm can learn better if it is allowed to choose the training data. AL algorithms can be uncertainty based, optimization based or Bayesian. The most common among these is the uncertainty based methods that rely on entropy (Joshi et al., 2009) or distance of data points from decision boundaries (Brinker, 2003; Tong and Koller, 2001). Bayesian methods use non-parametric models that are applicable to small datasets (Kapoor et al., 2007).

Despite renewed interest in CNNs in the recent years, AL for CNNs remains a largely unexplored territory. (Wang et al., 2016) proposed a Cost-Effective Active Learning (CEAL) framework for CNN based classifiers. CEAL combines samples classified with high confidence scores with the most informative samples annotated by humans in order to update the model. Unlike CEAL, we propose to render and generate synthetic images for model update, thereby eliminating human intervention. (Stark et al., 2015) proposed a method to actively train CNNs based on uncertainity scores and showed that CNNs can be trained iteratively for CAPTCHA recognition. Our work rather focuses on hard imaging conditions. The main challenge preventing CNNs from being tweaked for AL is that CNNs are batch based, requiring several training samples in a batch. In a recent unpublished work, (Sener and Savarese, 2017) show that the problem of selecting a batch of samples for CNNs can be addressed using a core-set approach both for supervised and semi-supervised learning. While all these focus on classification, we show active learning for object detection.

# 3 Imaging Conditions and Data Synthesis

Supervised learning algorithms rely heavily on annotated training data and CNNs are no exception. With images getting increasingly uploaded in sites such as Flickr, there is no shortage of data, even with in the wild imaging conditions. But these real world data come with two shortcomings. First, they come without annotations and so manual annotation is essential to put to use this vast gamut of data. Second, the manually annotated data will prove useful for object detection at category-level, but training to detect a particular instance of an object requires many more annotated images of the same object. We resort to generate synthetic data and overcome these shortcomings as explained in section 3.3.

## 3.1 Imaging Conditions

Before we developed our algorithm to detect deformable objects under hard imaging conditions, we did an extensive analysis on all possible imaging conditions in the wild. Our findings are summarized in table 1 and example images for conditions chosen for our algorithms are shown in figure 3. We have grouped them under five categories. Additionally, for each item under a category, we have listed the assumptions we make while detecting them alongside their corresponding parameters. For instance, it is fair to assume that the target object is at least partially visible under occlusion. General assumptions in our analysis include the perspective camera model and smooth and topology preserving deformations of objects. We now explain each of the items per category.

### 3.1.1 Scene

A scene is defined as that part of the world seen by the camera producing the image. Possible hard conditions in a scene could be any of:

***Self-occlusions.*** A highly deformed object can be occluding parts of itself. For instance, imagine a paper rolled tightly. Object texture cues occluded inside the roll are lost.

***Clutter.*** In a scene with many objects, the background tends to cause difficulty to detection. A cloth besides a magazine cover (the target object) with a texture similar to it can be detected along with the target, hampering the detection accuracy.

Table 1: Imaging conditions in the wild.

| Imaging conditions | Assumptions | Parameters |
| --- | --- | --- |
| *Scene* | | |
| Self-occlusion | object partially visible | level of occluded pixels: none, mild, strong |
| External occlusion | object partially visible | level of occluded pixels: none, mild, strong |
| Clutter | object partially visible | texture of clutter objects |
| *Object Surface Appearance* | | |
| Lighting | no full saturation, object texture visible | # sources, type of source position, intensity, light texture |
| Shadows | object partially visible | none |
| Specularities | no saturation of regions | none |
| Foldings or wrinkles | none | area, frequency and sharpness of wrinkles |
| *Imaging* | | |
| Camera pose | object partially visible | obliqueness, rotation, translation |
| FoV | object partially visible | percentage of occluded pixels: none, mild, strong |
| Focus blur | even blur on full image | blur radius (kernel size) |
| Motion blur | even blur on full image, linear motion | distance of motion in pixels, angle of motion |
| *Scale* | | |
| Object size | object detectable by a human | object size to image diagonal ratio |
| *Camera setting* | | |
| Intrinsics of the camera | mild distortion | focal length |

**External occlusions.** In a scene with many objects, an object may be overlaying on the target thereby making detection harder.

### 3.1.2 Object Surface Appearance

In spite of factors like pose changes, the same object may appear differently on different images. It is largely dictated by how the objects in a scene interact with the light sources:

**Lighting.** A scene can be lit by natural (sun) or artificial sources. Further, the number of sources, their location, intensity and texture (color) may change the way an object looks.

***Shadows.*** Shadows are cast by other objects that obstruct light falling on the target object. Pixels with shadows may bring information depending on the type of shadow (umbra, penumbra, antumbra).

***Specularities.*** Mirror-like reflections on the surface of an object are called specularities. If specularities are limited, they may hinder detection. However, if the reflections are saturated, then there is no directly usable information in the region.

***Foldings or wrinkles.*** Foldings or wrinkles bring unsmoothness to the surface. However, the difficulty caused to detection is quite limited because the overall change in appearance is limited.

### 3.1.3   Imaging

Imaging refers to how a given object is captured by a camera. We find below four possible imaging variations:

***Camera pose.*** The same object can be captured quite differently based on the location and orientation of the camera. It can be rotated and be at different locations from the object.

***Field of view (FoV).*** The camera may be very close to the object which may not even be fully covered by the FoV of the camera, in which case we consider the pixels occluded or lost.

***Focus blur.*** Optical or focus blur is caused by the object not being focused by the camera. We assume uniform focus blur.

***Motion blur.*** A hand-held camera imaging a scene undergoes minor shakes leading to motion blur in the image. It is much exaggerated in an image captured by long exposure. We assume uniform motion blur in the image.

### 3.1.4   Scale

The scale of a given deformable object can vary significantly. For instance, a newspaper in an outdoor cluttered scene may be very small compared to being on a table indoors. The detection of objects at different scales brings a level of challenge despite pose changes and other factors.

### 3.1.5   Camera Settings

The camera used for capturing the image may not be ideal. For instance, there can be lens distortions and additive noise due to image processing by the camera hardware to store the image captured. We assume mild lens distortion in our setting.

Table 2: Quantitative results for all imaging conditions considered for Fast R-CNN. In order to short-list conditions out of all the conditions analysed, we tested for each of the conditions with a model trained with canonical conditions.

| Imaging conditions | mAP scores |
|---|---|
| Focus blur | 0.3377 |
| Motion blur | 0.4577 |
| Pose change | 0.4352 |
| Deformation | 0.5644 |
| Scale | 0.5947 |
| Occlusion | 0.5724 |
| Clutter | 0.9521 |
| Lighting | 0.6681 |
| Shadows | 0.5887 |
| Specularities | 0.8110 |
| Wrinkles | 0.9434 |
| Noise | 0.8349 |

Table 3: Range of values uniformly sampled for each difficulty level and imaging conditions.

| Conditions | Parameters | Units | Difficulties and values | | | |
|---|---|---|---|---|---|---|
| | | | Easy | Medium | Hard | Canonical |
| Focus blur | kernel size | percents of image size | [0.39, 0.97] | [1.17, 1.95] | [2.14, 2.92] | 0 |
| Motion blur | linear motion | percents of image size | [0.39, 0.97] | [1.17, 2.92] | [3.12, 4.88] | 0 |
| | angle | radians | $[0, \pi]$ | $[0, \pi]$ | $[0, \pi]$ | 0 |
| Pose change | roll, pitch & yaw | radians | $[0, \pi/6]$ | $[\pi/6, \pi/3]$ | $[\pi/3, \pi/2]$ | 0 |
| | position | meters | [0, 0.5] | [0.5, 1] | [1, 2] | 0 |
| Deformation | # rulings | - | 3 | 5 | 8 | 2 |
| Scale | small | proportion of the object | [0.75, 1] | [0.5, 0.75] | [0.1, 0.5] | 1.0 |
| | large | proportion of the object | [1, 1.5] | [1.5, 2.25] | [2.25, 3] | 1.0 |
| Occlusions | visibility | proportion of object visible | [0.7, 0.9] | [0.3, 0.7] | [0.1, 0.3] | 1.0 |
| Lighting | irradiance (dark) | watt/$m^2$ | [0.5, 1.0] | [0.05, 0.5] | [0.01, 0.05] | 1.0 |
| | irradiance (bright) | watt/$m^2$ | [0.5, 1.0] | [1.0, 5.0] | [5.0, 10.0] | 1.0 |

## 3.2  Chosen Conditions and Parameters

Our initial curiosity to study the behavior of CNN based detection algorithms under hard imaging conditions led us to experimentally probe the detection algorithms for all the conditions listed in table 1. For this, we rendered 500 canonical images per object thereby generating a training set of 2,500 images. The Fast R-CNN detection algorithm was initialized with an ImageNet pre-trained model and trained with these images for 20,000 iterations. The trained model was tested for each of the conditions listed in table 1. For the test set, we generate 330 test images for each difficulty, making a test set of 990 images for each condition.

The quantitative results of our findings are listed in table 2. Wrinkles, specularities, noise and clutter do not seem to impact CNN based detection algorithms significantly. We have short-listed conditions that clearly degrade the detection performance in table 3. For all our experiments we consider only these conditions and ignore the rest of the conditions from table 1.

Shadows form a common and important phenomenon. However, they do not need a specific treatment in our analysis and can simply be handled as occlusions. The reason is that an occlusion corrupts the objects observed colour to spatially varying values, while a shadow dims the colours, generally leading to a dark, almost black area. In terms of the image observations, shadows thus form a special case of occlusions.

Alongside the short-listed conditions are parameters that define the conditions and the range of values we chose for each. Additionally, we grouped the ranges into three difficulty levels: *easy*, *medium* and *hard*, both for the sake of training and testing. For every new image rendered for a given condition and difficulty, we stochastically sample in these ranges to obtain unique values. For instance, the size of the kernel used for focus blur with difficulty level easy could range from 0.39% to 0.97% of the image size. We now explain the parameters for each condition and how we use them in our rendered images. Figure 3 shows examples of rendered images for each of the imaging conditions and the corresponding difficulties.

**Focus blur.** We performed uniform Gaussian blur by choosing a blur kernel in the difficulty range specified. We did this after adding the background to the rendered images.

**Motion blur.** We assumed uniform linear motion and chose parameters corresponding to linear motion in pixel units and the angle of motion in degrees. We induced motion blur after adding background to the rendered images.

**Pose change.** Camera pose is defined by the roll, pitch and yaw Euler angles and its position. Position changes (translation) are in meters from the initial camera position during object rendering.

**Deformations.** Deformation parameters correspond to the number of guiding rulings in the developable surface and the number of regions into which the surface is divided (Perriollat and Bartoli, 2013). The larger the number of rulings, the stronger the deformation and so the more difficult the detection.

***External occlusion.*** We masked the rendered image based on the size and location parameters and added external occlusions accordingly. The size of the occluding object is the proportion of the target object. Location was chosen randomly but within the target object mask before adding the background.

***Scale.*** An object can be rendered smaller or larger than the size of its mesh. We chose both as multiples of the original size.

## 3.3 Data Synthesis

Given that we tackle deformable objects, we endeavor to generate objects with several random deformations. Several types of deformation exist. We focus on developable surfaces, that undergo isometric deformations. Many deformable objects as well as rigid objects are isometric.

We generate random deformations of isometric surfaces using the paper model toolbox. Example 3D meshes generated for a given set of parameters are shown in figure 1. We import the 3D mesh into the powerful rendering tool Blender. Blender provides a way to seamlessly change ambient lighting, pose and scale of the imported 3D object. We exploit this possibility to render all possible variations in imaging conditions to the 3D mesh. A schematic of the rendering pipeline is shown in figure 2. The randomly deformed meshes from the paper model toolbox are rendered with the target texture maps. A randomly chosen background image is added to the rendered image to generate training or test images along with bounding box annotation.

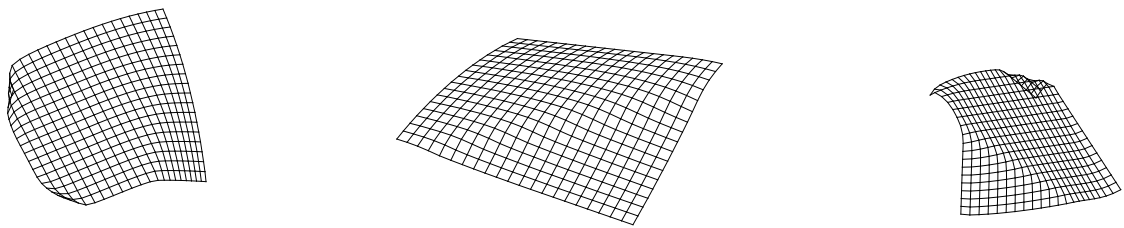

Figure 1: Meshes generated with random deformations using the paper model toolbox with 3 rulings and 3 regions.

## 4 Active Learning Algorithm

Having described the method to automatically generate synthetic images and their bounding box annotations on the go, we now introduce our active learning algorithm that can be leveraged to detect deformable objects in many hard imaging conditions.

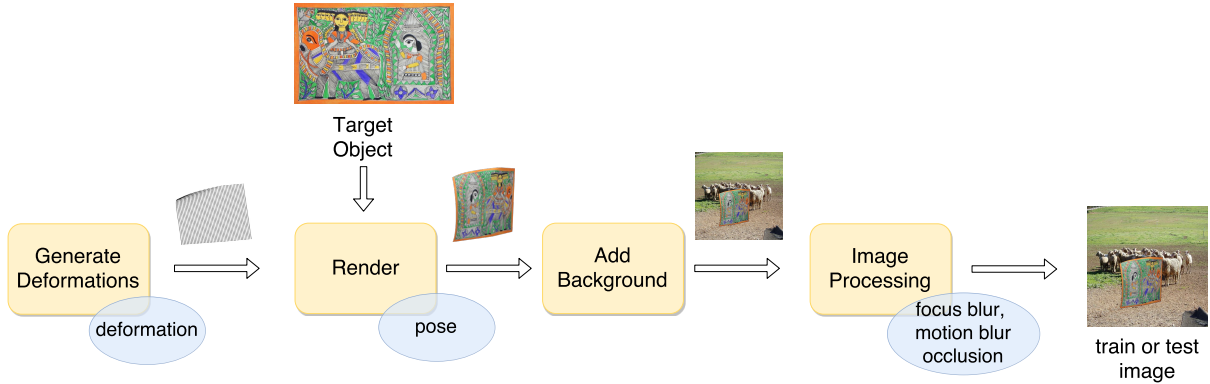

Figure 2: The proposed rendering pipeline which allows us to control the imaging condition parameters of table 1.

## 4.1 Notation

We first summarize the notation used in our algorithm in table 4 before delving into the nitty-gritties.

## 4.2 Proposed Algorithm

Since the introduction of R-CNN (Girshick et al., 2016) publications that evaluated their algorithms on the PASCAL VOC dataset in the past couple of years have relied heavily on transfer learning rather than learning from scratch by random initialization of weights. In this paradigm, the weights are initialized with the weights from models pre-trained on ImageNet rather than with random values. This approach, commonly dubbed *fine-tuning*, has been shown to perform superiorly compared to initializing with random weights (Agrawal et al., 2014). We leverage this idea profitably to *actively generate and learn* from examples of hard imaging conditions as and when our models fail for those.

**Algorithm.** The algorithm we propose is shown in Algorithm 1. For the sake of simplicity, we have ignored the superscripts from our notation. For instance, $\omega_1$ stands for $\omega_1^{c,\delta}$. As mentioned in section 3.2, we break down the different imaging conditions into three levels of difficulty and iteratively learn one at a time. We further break down the iteration of a given condition and difficulty into $k$ increments of training. This brings two features. First, the mAP scores evaluated for each of the $k$ iterations help introduce the stopping criteria, which makes our algorithm an active learning algorithm as we will explain next. Second, the algorithm learns from very few images (as little as 25) without overfitting to a particular condition and difficulty under consideration at the same time. The resulting model of each iteration $k$, initializes the weights while training in iteration $k+1$.

**Stopping criterion** During each training iteration $k$ we obtain a model $\omega_k$. If we choose a sequence of a fixed number of models and their mAP scores, we call this a *window*. We choose two such windows: window $M$ of the most recent models and window $N$ of models just prior to window $M$. We choose to stop the increments of training $k$ and move on to the next condition or difficulty if $|S_M - S_N| \leqslant \tau$ where $\tau$ is a threshold and $S_M$ and $S_N$ are the max values of mAP scores in the corresponding windows. We call this the *stopping criterion*. $\omega_{be}$ is the model corresponding to the best mAP score out of $S_M$ and $S_N$, which initializes the weights to train for the next condition or difficulty.

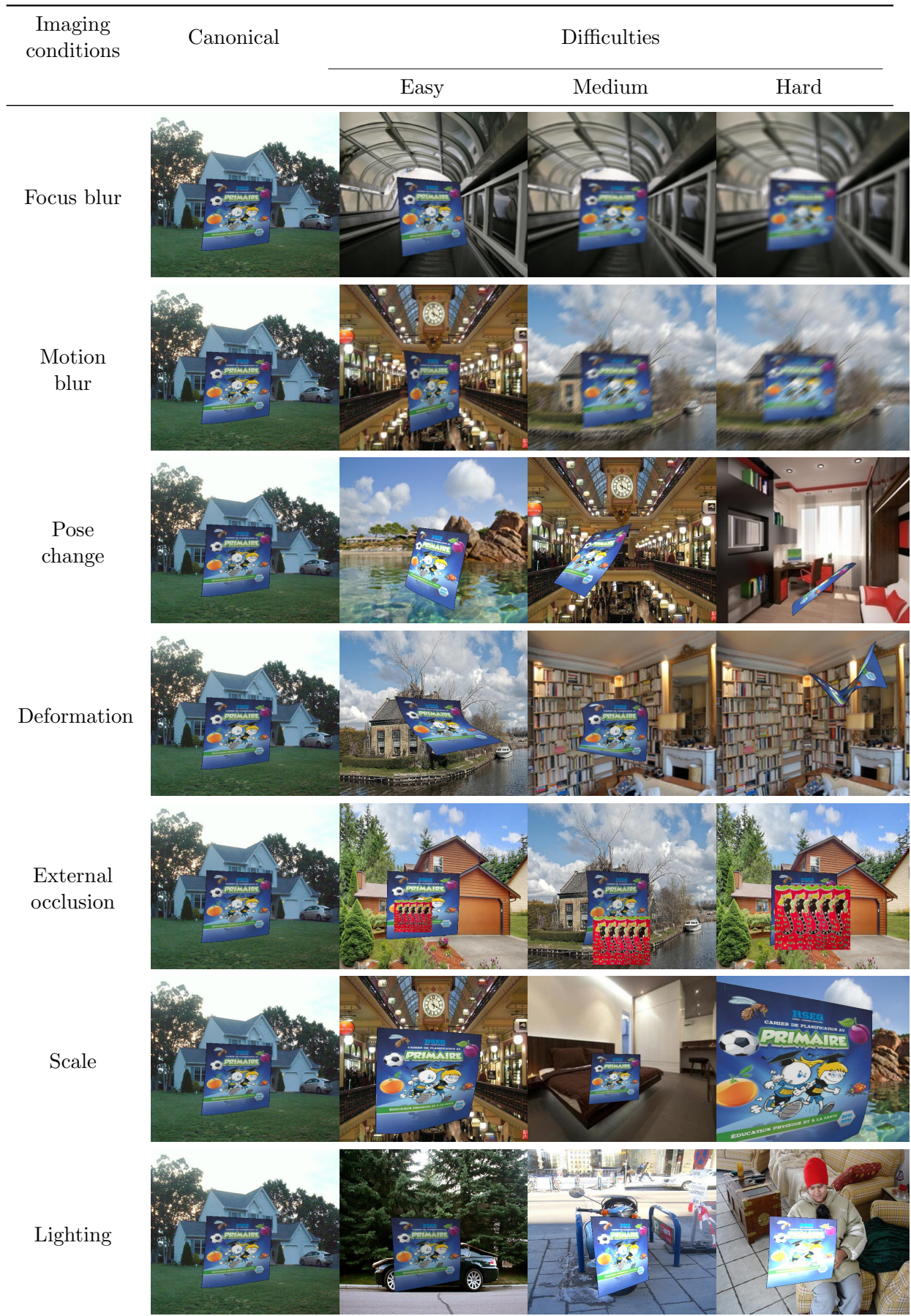

Figure 3: Images generated for different imaging conditions and difficulty levels, with random background.

Table 4: Notation used in the proposed algorithms.

| | |
|---|---|
| $C = \{\text{fb, mb, po, de, eo, sc, li}\}$ | Set of all imaging conditions (focus blur, motion blur, pose change, deformations, external occlusion, object scale, lighting) |
| $\Delta^c = \{\delta_e^c, \delta_m^c, \delta_h^c\}$ for $c \in C$ | Set of difficulties (easy, medium, hard) for a given condition $c$ |
| $k$ | Index for training increments during active learning |
| $\omega$ | A model |
| $\Omega^{c,\delta} = \{\omega_1^{c,\delta}, \omega_2^{c,\delta}, \ldots, \omega_k^{c,\delta}\}$ | Set of trained models for all iterations of active learning $k$ |
| $\omega_{be}$ | Best model chosen when the stopping criterion is met |
| $S_{tr}$ | Set of training images and their bounding boxes $\{I_{tr}, B_{tr}\}$ |
| $S_{te}$ | Set of test images and ground truth bounding boxes $\{I_{te}, B_{te}\}$ |
| $B_{pr}$ | Bounding boxes predicted by the object detector |
| $\theta_i^{c,\delta}$ | Stochastically chosen parameters for a given image $i$ |
| $n, m$ | Number of test and train images respectively in an iteration |
| $\Theta_n^{c,\delta}, \Theta_m^{c,\delta}$ | Set of all $\theta$ values, $\{\theta_1, \theta_2 \cdots \theta_n\}, \{\theta_1, \theta_2 \cdots \theta_m\}$ |
| $P$ | Set of mAP scores for all models in $\Omega$ |
| $t$ | Training iterations of the detection algorithm |
| $generate\_parameters$ | Gives a set of parameters to render images for a given condition and difficulty level |
| $generate\_data$ | Renders and generates dataset using the given parameters |
| $train\_detector$ | Given the training set and model for initialization, trains the detection algorithm |
| $detect\_object$ | Detects target objects given the trained model and test dataset |
| $evaluate\_predictions$ | Given the prediction results and ground truth, returns the mAP scores |

A problem in which the learner has the ability to influence and select its training data is termed an *active learning* problem (Cohn et al., 1996). The introduction of the stopping criterion in our algorithm makes it learn actively. The algorithm decides wether more training examples are needed for a given imaging condition and difficulty or if it has learnt it successfully to move on to the next condition or difficulty.

### 4.3  Learnability of Imaging Conditions

In our algorithm, the underlying assumption was that imaging conditions can be learned by showing examples. In this section, we get down to analyzing how well each imaging condition can be learned by the baseline detection algorithms.

Algorithm 2 shows the pseudo code for analyzing the learnability of imaging conditions. Here, for each condition, we generate $n$ test images per difficulty level and keep these as a fixed test set during active learning iterations. We iteratively train for each difficulty level under the condition. Because we are interested in comparing the performance for each condition, we generate a test set for only that condition. Furthermore,

---

**Algorithm 1:** Proposed Active Learning Algorithm

**Data**: Imaging conditions $C$
  Difficulty levels for imaging conditions $\Delta$
  Initial model $\omega_{in}$
  Size of the most recent window of models $M$
  Size of the models just before the window $N$
  Number of training images for each difficulty under each condition $m$
  Number of test images for each difficulty under each condition $n$
**Result**: Final model $\omega_{be}$
// generate a standard test dataset
$\Theta_{te} \leftarrow \varnothing$
**for** $c \in C$ **do**
  **for** $\delta \in \Delta$ **do**
    $\Theta_{te} \leftarrow \Theta_{te} \cup \{generate\_parameters(c, \delta, n)\}$
$S_{te} \leftarrow generate\_data(\Theta_{te})$
**for** $c \in C$ **do**
  **for** $\delta \in \Delta^c$ **do**
    $k \leftarrow 1$
    $stop \leftarrow false$
    $\Omega \leftarrow \{\omega_{in}\}$
    $P \leftarrow \varnothing$
    **while** *not stop* **do**
      // generate training set
      $\Theta_{tr} \leftarrow generate\_parameters(c, \delta, m)$
      $S_{tr} \leftarrow generate\_data(\Theta_{tr})$
      $\omega_k \leftarrow train\_detector(S_{tr}, \omega_{in})$
      $\Omega \leftarrow \Omega \cup \{\omega_k\}$
      $B_{pr} \leftarrow detect\_object(\omega_k, I_{te})$
      $P \leftarrow P \cup \{evaluate\_predictions(B_{pr}, B_{te})\}$
      **if** $k \geq M + N$ **then**
        $P_M \leftarrow max(P(end - M + 1, end))$
        $P_N \leftarrow max(P(end - (M + N) + 1, end - M))$
        **if** $|P_M - P_N| \leqslant \tau$ **then**
          // plateau found
          $\omega_{be} \leftarrow \Omega(argmax(P))$
          $\omega_{in} \leftarrow \omega_{be}$
          $stop \leftarrow true$
      **else**
        $\omega_{in} \leftarrow \omega_k$
      $k \leftarrow k + 1$

---

we disregard the stopping criterion set for active learning. We rather fix the number of iterations $k$ to a constant, $\kappa$. This is again to ensure a fair comparison. Because the training iterations $t$ is fixed, by fixing $k$, we ensure each difficulty level $\delta$ under a condition $c$ is trained only for $kt$ iterations. For instance, if $k$ is chosen as 10 and $t$ as 1,000, images generated with easy level of focus blur will be trained for 10,000 iterations and so do all other difficulties under each imaging condition. Our experiments and findings on learnability are presented in section 5.2.

## 5   Experiments and Results

We report our experimental results using both synthetic and real images. All experiments were conducted using synthetic images unless otherwise specified. All results are reported using the standard mAP metric. None of the objects considered are part of the ImageNet dataset. However, all the detection algorithms

---

**Algorithm 2:** Learnability of imaging conditions

**Data**: Imaging conditions $C$
Difficulty levels for imaging conditions $\Delta$
Initial model $\omega_{in}$
Number of training images for each difficulty under each condition $m$
Number of test images for each difficulty under each condition $n$
Number of incremental learning iterations for a given condition and difficulty $\kappa$

**Result**: mAP Scores $P$

$S_{tr} \leftarrow \varnothing$;
$\Omega \leftarrow \omega_{in}$;
**for** $c \in C$ **do**
$\quad S_{te} \leftarrow \varnothing$;
$\quad$ **for** $\delta \in \Delta$ **do**
$\quad\quad \Theta_{te} \leftarrow generate\_parameters(c, \delta, n)$;
$\quad\quad S_{te} \leftarrow S_{te} \cup generate\_images(\Theta_{te})$;
$\quad$ **for** $\delta \in \Delta$ **do**
$\quad\quad k \leftarrow 1$;
$\quad\quad$ **while** $k \leq \kappa$ **do**
$\quad\quad\quad \Theta_{tr} \leftarrow generate\_parameters(c, \delta, m)$;
$\quad\quad\quad S_{tr} \leftarrow generate\_images(\Theta_{tr})$;
$\quad\quad\quad \omega_k \leftarrow train\_detector(S_{tr}, \omega_{in})$;
$\quad\quad\quad \Omega \leftarrow \Omega \cup \omega_k$;
$\quad\quad\quad B_{pr} \leftarrow detect\_object(\omega_k, I_{te})$;
$\quad\quad\quad P \leftarrow P \cup \{evaluate\_predictions(B_{pr}, B_{te})\}$
$\quad\quad\quad \omega_{in} \leftarrow \omega_k$
$\quad\quad\quad k \leftarrow k + 1$

---

considered in our experiments use ImageNet pre-trained models for initialization. In all the detection algorithms, the last layer of the network was modified to output the confindence values for 5 objects. Similarly, the detection coordinates output by the last layer of the network were modified to be just for 5 objects instead of 20 corresponding to the PASCAL VOC dataset. Also note that training for the easy level of difficulty can be seen as fine-tuning the ImageNet models for our objects.

## 5.1 Dataset

The objects chosen for our experiments are the textures extracted from the covers of randomly chosen magazines on the web. We chose these as magazine covers are truly developable surfaces. They can be printed on a paper and deformed easily by hand for real world experiments as shown in section 5.5.2.

As explained in previous sections, there are seven conditions considered in our study. We specified the range for parameters corresponding to each in table 3. When we render images with parameter values lesser than the *easy* level of difficulty, we say these images have *canonical conditions*. More specifically, these images have no focus or motion blur. The pose of the camera is fixed facing the rendered object with its scale unchanged. The object is deformed with 2 guiding rulings and 3 regions. No external occlusions are added. All experiments were conducted with 30 images per difficulty under each imaging condition. We chose to have 5 objects in our database. The 7 conditions and 3 difficulty levels make a total of 3,150 images. We refer to this as the *standard test set*. When an image is generated with a particular difficulty and condition, the other conditions are kept to canonical values. For instance, images with motion blur do not have variation in pose or deformation. Rather, they are set to canonical values. The training set size and conditions are explained in each section. While rendering external occlusions, we randomly choose an occluding object out of 50 other objects. Similarly, while adding backgrounds, for each rendered image we randomly chose one from 17,100 in the PASCAL VOC 2012 dataset.

All experiments were conducted with the detection confidence threshold for the object detection algorithms set to 0.8. The Intersection over Union (IoU) score for evaluation was set to 0.5. For the stopping criterion, we set the window sizes, $M = 3$ and $N = 8$ and a threshold value of $\tau = 0.02$.

## 5.2  Learnability of Imaging Conditions

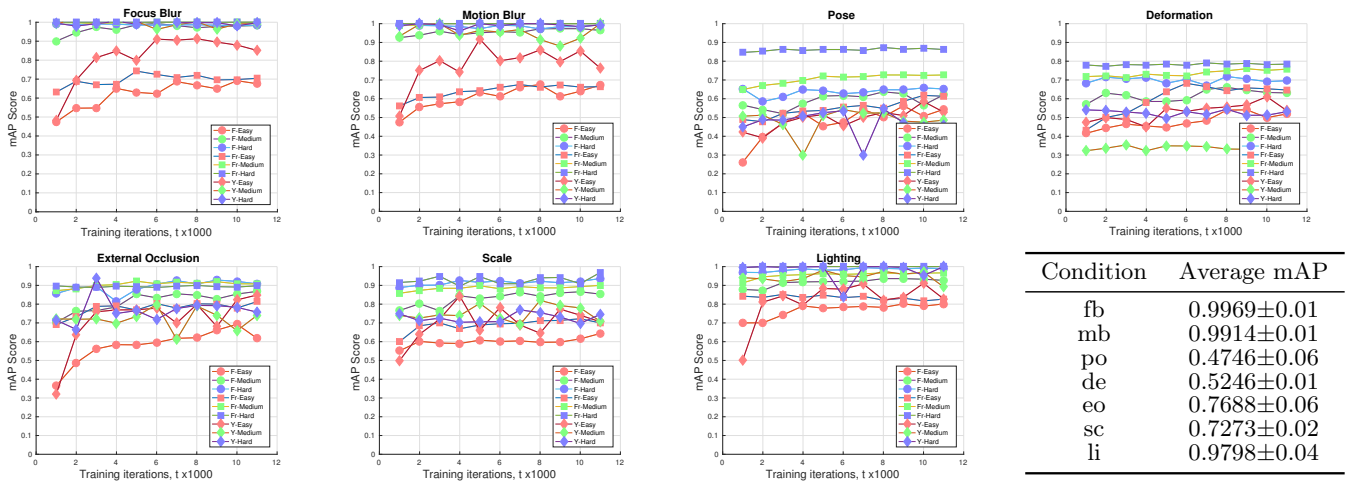

Figure 4: Learnability of each condition. In each case we can notice that training with a higher level of difficulty helps improve the mAP scores. Here F stands for Fast R-CNN, FR for Faster R-CNN and Y for YOLO9000.

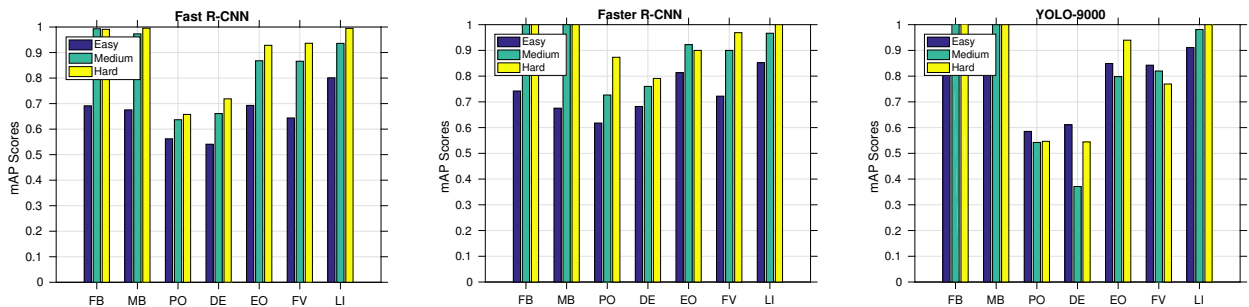

Figure 5: Best mAP scores for learnability of imaging conditions. Pose and deformations are amongst the most difficult to learn by CNN based detection algorithms. Nevertheless, they prove to be learnable provided sufficient examples are used for training as we see the mAP scores going up when we train with hard cases.

In our active learning algorithm, we learn one condition at a time. So we have the possibility to examine how well the CNN copes with each condition separately. Here, we do just that and show the results corresponding to algorithm 2.

For the test set, we generate 330 test images for each difficulty, making a test set of 990 images for each condition. While training, we use 25 images per iteration $k$ and retaining them in the following iterations. Each of the detectors tested was initialized with the pre-trained models available for them. More specifically, we initialize Fast R-CNN with CaffeNet, Faster R-CNN with VGG-Net and YOLO9000 with the pre-trained weights released with the code.

The results for each condition per iteration for all the object detectors compared are shown in figure 4 and their best mAP scores per condition are shown in figure 5. Interestingly, both focus blur and motion blur can be learned very well by the networks. In both cases, training with an easy level of difficulty did not make much difference but there is a seismic shift in performance as soon as we train for a medium level of blur. Pose changes and deformations prove to be the most challenging. Given that all the object detectors show similar behaviour for learnability indicates that the limitation is not dependent on the detection algorithm but is inherent to CNNs. In summary, the most learnable to the least learnable based on our findings is as follows: focus blur, motion blur, scale changes, occlusions, deformations and pose changes. YOLO9000 at times exhibits minor dip in mAP scores in the bar chart with increasing difficulty. But this can be attributed to the volatility of the mAP scores while training YOLO9000. Overall we note that Faster R-CNN performs best for learnability.

## 5.3  Resetting the Training Dataset

In algorithm 1 the training set $S_{tr}$ is reset for every active learning iteration $k$. In order to arrive at this decision, we experimented with both resetting and incrementing $S_{tr}$. Algorithm 1 implements the resetting case where $S_{tr} \leftarrow generate\_data(\theta_{tr})$. In order to increment $S_{tr}$, we used the modified algorithm 1 such that $S_{tr} \leftarrow S_{tr} \cup generate\_data(\theta_{tr})$. Our motivation is to study the gain we obtain by reusing the past images already used for training. The results of comparison for a randomly chosen imaging condition are shown in figure 6 for all three object detectors considered.

We notice that the mAP scores at each iteration $k$ follow a similar trend in both the cases. We also noted that the training time with incrementing was significatly higher than that of resetting the training set. With these results we concluded that we will reset the training set for each iteration $k$ in our active learning algorithm.

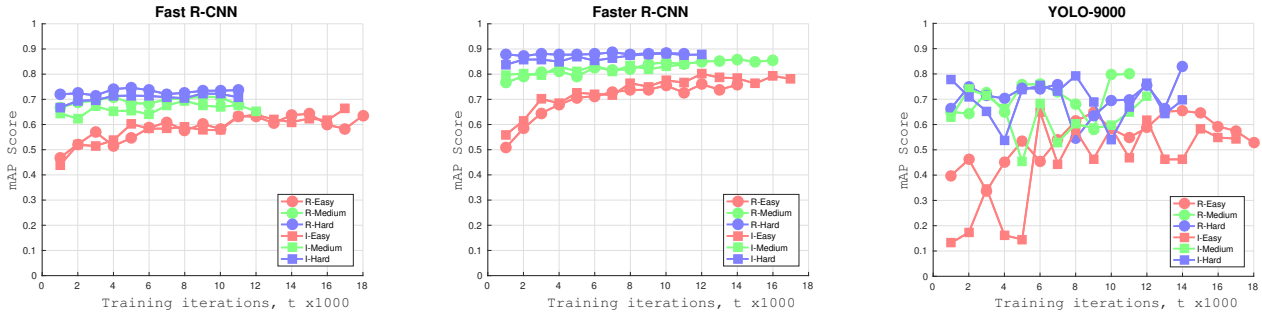

Figure 6: A comparison of refreshing training set against retaining the training set while training for deformations shows similar performance indicating the ability of CNNs to retain their learning as $k$ increases. Here R stands for refreshing the training set and I for incrementing the training set.

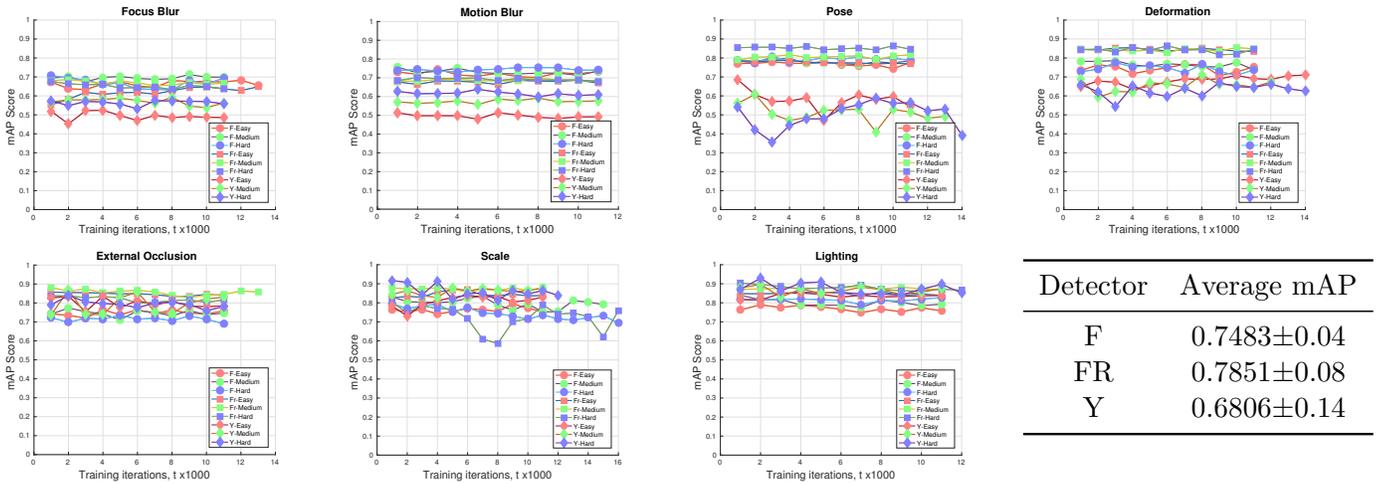

Figure 7: Results of running the active learning algorithm on three state-of-the-art detection algorithms. Here F stands for Fast R-CNN, FR for Faster R-CNN and Y for YOLO9000.

## 5.4 Active Learning

The findings presented in this section correspond to running Algorithm 1 using three state-of-the-art object detectors, namely Fast R-CNN, Faster R-CNN and YOLO9000. For training, at each iteration of the incremental training $k$ we generated 10 images for each of the 5 objects considered for a given condition $c$ and difficulty $\delta$. Our stopping criterion was used to exit the active learning phase with the next condition being initialized with the best model from the current iteration. The training iterations $t$ for each increment $k$ was fixed to 1,000 for all the object detectors compared. Given that the stopping criterion pitches in after $k = 11$, the total number of training iterations for each combination of $\delta$ and $c$ would amount to at least 11,000.

The results of testing with the standard test set are shown in figure 7. The results are presented in the order in which the conditions were considered. We start with focus blur and ended with scale changes. We noticed that the order in which we choose the imaging conditions did not have any impact on the test performance. It shows an increasing trend in the mAP scores as we train for more conditions. This indicates that object detectors based on deep architecures can seamlessly tune their parameters to imaging conditions seen long before rather than forgetting them and tuning their parameters to recently seen conditions. The same behaviour can be observerd for all the detectors compared. The best mAP scores corresponding to the best models $\omega_{be}$ achieved for each imaging condition are shown in figure 8.

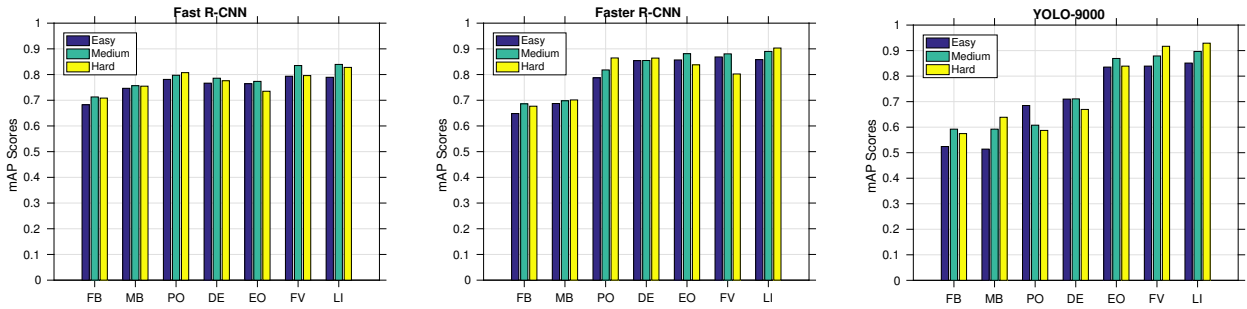

Figure 8: mAP scores corresponding to the best models $\omega_{be}$ for the active learning algorithm whilst learning the imaging conditions.

## 5.5 Detection Results

We show the result of running our active learning algorithm on both synthetic and real images. We show both positive and negative cases.

### 5.5.1 Synthetic Images

Using the final model we obtained with the active learning algorithm, we show object detection results on some synthetic images. As shown in figure 9, though we are testing with our final model, the model can detect objects under the range of conditions being trained for including blur, deformations and pose changes. Figure 10 shows cases of false negatives. Some conditions like deformation and scale in the figure look difficult even for the human eye.

### 5.5.2 Real Images

The performance of a model trained on synthetic images tends to degrade when tested on real images. This difference in performance is due to *domain adaptation* (Ben-David et al., 2010; Crammer et al., 2008). When training Machine Learning algorithms, the training data is assumed to be drawn from some fixed *source* distribution. If the test data is from a different *target* distribution it requires domain adaptation. In domain adaptation, both the source and target domain images have the same objects. Though the problem of domain adaptation is beyond the scope of this work, we show that models trained actively on synthetic data work well on real data by training with as little as 100 real images.

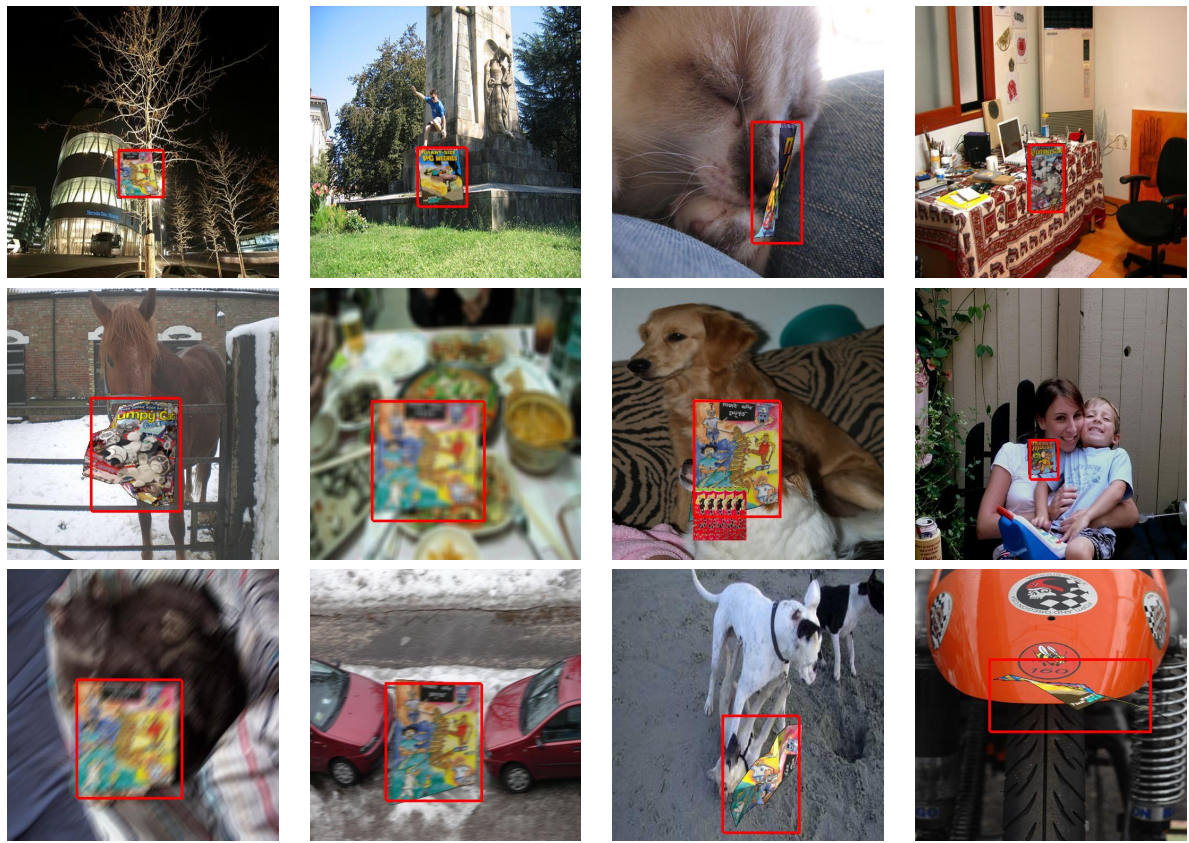

Figure 9: Selected examples of detection results on synthetic data using the proposed active learning algorithm. Our final model can cope with the range of imaging conditions trained for during training.

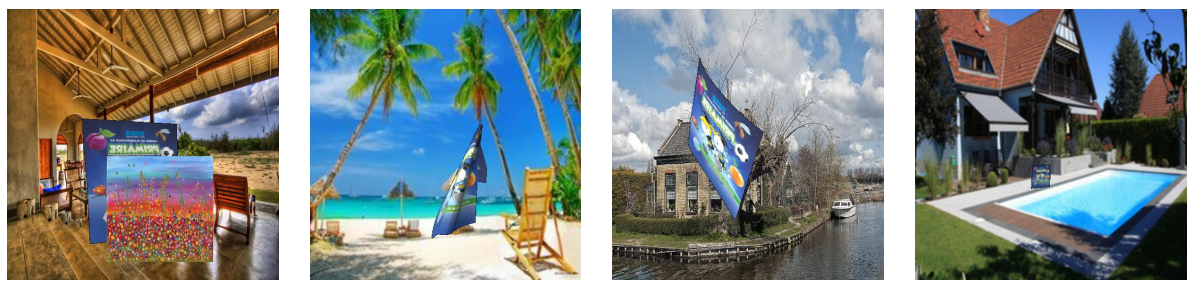

Figure 10: False negative cases by running the proposed active learning algorithm on synthetic images. From left to right: occlusions, deformations, pose and scale. As can be seen, some of the conditions like deformations and scale are hard even for the human eye.

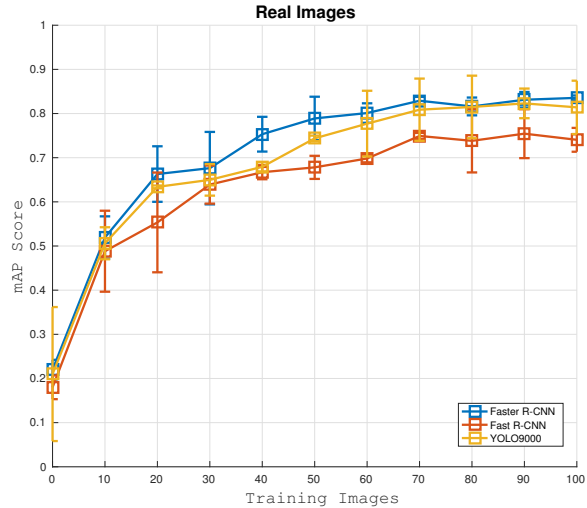

Figure 11: Quantitative results on our annotated dataset of real images indicate that our algorithm can seamlessly cope with hard imaging conditions in real world scenarios by training with as little as 100 real images

To test our algorithm qualitatively and quantitatively on real data, we captured and annotated two datasets (indoors and outdoors) with the objects undergoing different imaging conditions such as blur, pose and deformations. After pruning irrelevant frames, the final dataset consists of 1,285 images. The final model obtained by the active learning algorithm using synthetic images $\omega_{be}$ was used for fine-tuning on these real images. The quantitative results of different train-test split for this dataset are presented in figure 11 for all detectors considered. With mAP scores reaching a plateau with as little as 100 real images, we noticed that there is no need for an active learning approach to generalize on real images. In just 1,000 training iterations we could reach the performance on par with synthetic images. However, training on real images initialized with ImageNet pre-trained models did not converge after 1,000 training iterations. The results corresponding to the number of training images of 0 is for the test done on the model trained with synthetic dataset alone.

In figure 12, we show the positive results of testing our final model on some real images. We can see results similar to those of synthetic images. An interesting finding is that the final model works well on images with a combination of imaging conditions, indicating that an algorithm to train for a combination of imaging conditions in training images is unnecessary. Some false negative results are shown in figure 13. A cluttered environment easily produces false positives as shown at the bottom right image in the figure. We believe the general performance on real images can be improved by training with a larger pool of synthetic training images during incremental training. Also, adding a small proportion of real images annotated with bounding boxes to the synthetic images could improve the performance.

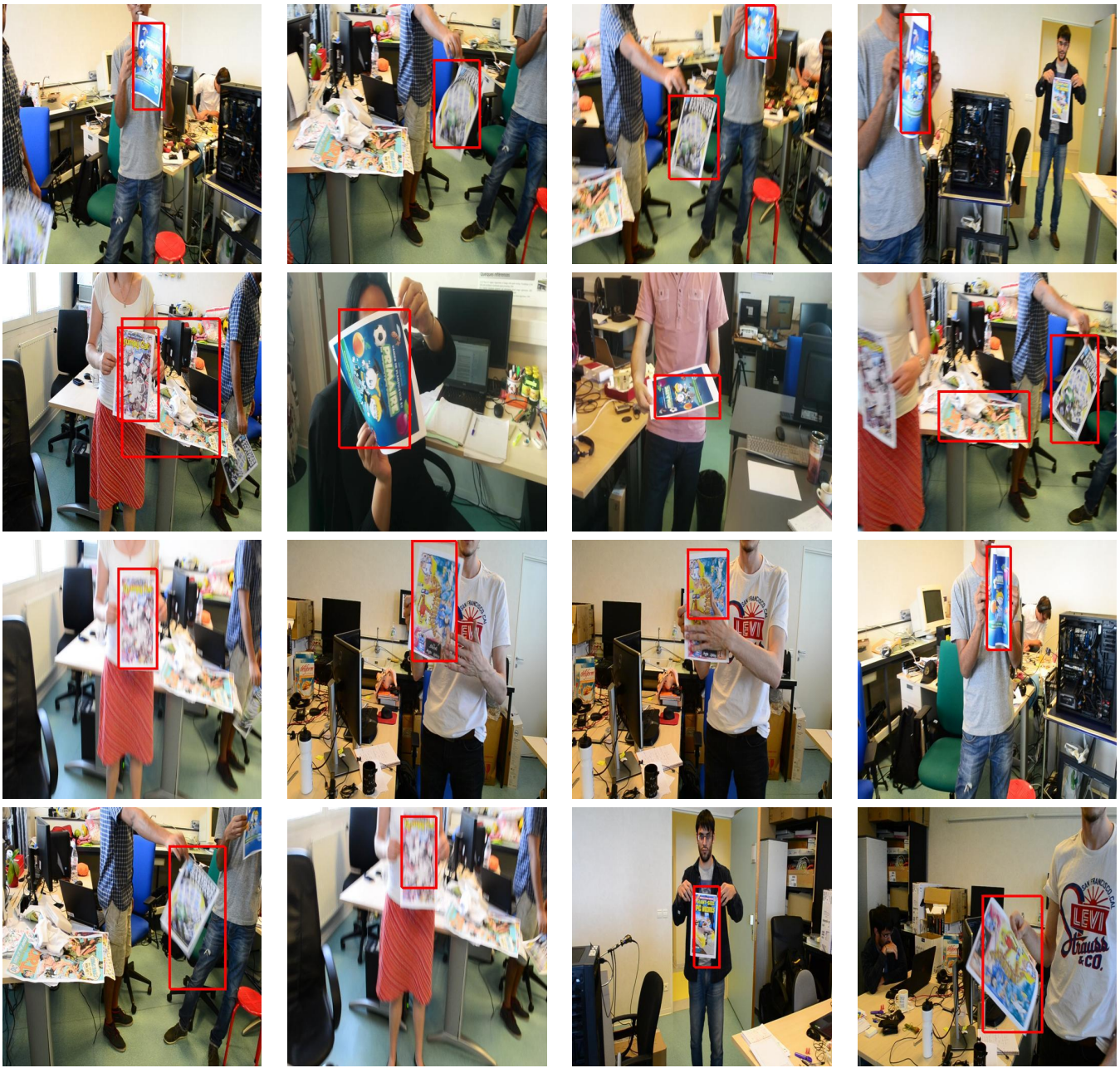

Figure 12: Selected examples of true positives on real images. The first image in the top row indicates that CNNs can cope naturally with lightings. The second image in the same row has a combination of motion blur and deformations. But the model still detects it.

## 6   Conclusion

We have proposed an active learning algorithm for CNN based object detection methods to detect deformable objects under hard imaging conditions. We studied the possibility of using synthetic data to train three state-of-the-art CNN based object detection algorithms namely Fast R-CNN, Faster R-CNN and YOLO9000 with a view to generalize on real images. We showed that leveraging the free annotations that come with synthetic data can be quite handy to train CNN architecture for domains where real images are scarce.

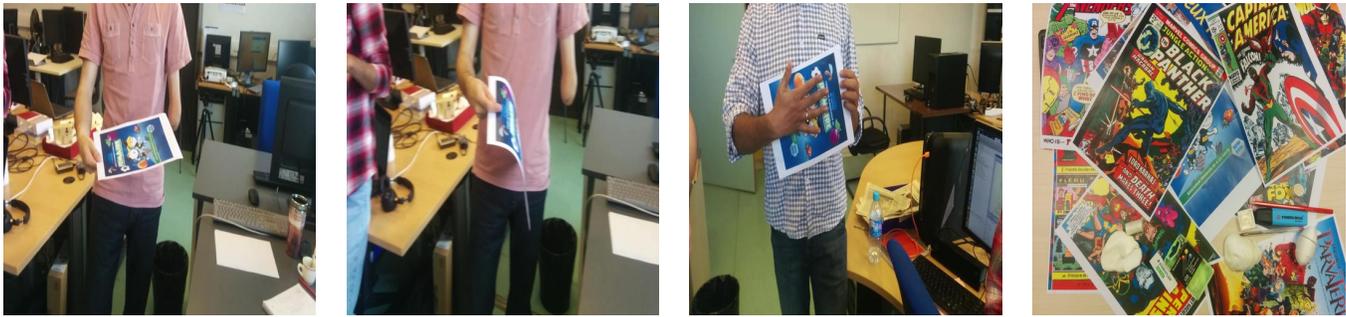

Figure 13: Examples of false negatives on real images.

We showed that different imaging conditions are learned differently by CNNs with pose and deformations amongst the hardest conditions to train a CNN with examples. Results on real data indicate the strength of using rendering to enrich the training dataset. Our experimental results prompt for a whole new approach to train CNNs and indicate a promising research direction.

As future work we propose to train the active learning algorithm for more than one condition and study its ability to learn more than one condition in a training iteration.